\documentclass[letterpaper, 10pt, conference]{ieeeconf}      

\IEEEoverridecommandlockouts                              

\usepackage{graphicx} 
\usepackage{amssymb}  
\usepackage[caption=false]{subfig}
\usepackage{multicol}
\usepackage{booktabs}

\usepackage[table,xcdraw]{xcolor}

\title{\LARGE \bf
Multi-view Incremental Segmentation of 3D Point Clouds for Mobile Robots
}

\author{Jingdao Chen$^{1}$, Yong K. Cho$^{2}$, and Zsolt Kira$^{1}$
\thanks{$^{1}$ Institute for Robotics and Intelligent Machines, Georgia Institute of Technology, Atlanta, GA 30332, USA
        {\tt\small jchen490@gatech.edu,zkira@gatech.edu}}
        \thanks{$^{2}$ School of Civil and Environmental Engineering, Georgia Institute of Technology, Atlanta, GA 30332, USA
                {\tt\small yong.cho@ce.gatech.edu}}
                }

\begin{document}

\maketitle
\thispagestyle{empty}
\pagestyle{empty}

\begin{abstract}
Mobile robots need to create high-definition 3D maps of the environment for applications such as remote surveillance and infrastructure mapping. 
Accurate semantic processing of the acquired 3D point cloud is critical for allowing the robot to obtain a high-level understanding of the surrounding objects and perform context-aware decision making.
Existing techniques for point cloud semantic segmentation are mostly applied on a single-frame or offline basis, with no way to integrate the segmentation results over time.
This paper proposes an online method for mobile robots to incrementally build a semantically-rich 3D point cloud of the environment. The proposed deep neural network, MCPNet, is trained to predict class labels and object instance labels for each point in the scanned point cloud in an incremental fashion. A multi-view context pooling (MCP) operator is used to combine point features obtained from multiple viewpoints to improve the classification accuracy. The proposed architecture was trained and evaluated on ray-traced scans derived from the Stanford 3D Indoor Spaces dataset. Results show that the proposed approach led to 15\% improvement in point-wise accuracy and 7\% improvement in NMI compared to the next best online method, with only a 6\% drop in accuracy compared to the PointNet-based offline approach.
\end{abstract}

\section{INTRODUCTION}

Mobile robots are frequently used to autonomously explore and survey unknown areas. To understand its environment, a mobile robot can use sensors such as cameras \cite{posada18}, depth sensors \cite{marion18}, and laser scanners \cite{antonello18} to perceive its surroundings. Regardless of the hardware setup, the acquired data can be summarized in point cloud form, which is an array of 3D points containing geometric and optionally color information. 
Recent studies~\cite{dai18}\cite{chencho18} have increasingly explored the use of 3D point clouds as a richer and more versatile representation of the environment.
Retrieving semantic information from point clouds is valuable for the robot to perform tasks such as obstacle detection, place recognition, and object retrieval.

However, current methods for point cloud semantic segmentation such as PointNet \cite{qi2017} and SGPN \cite{wangweiyue18} are lacking when applied to robotic real-time scanning because they are designed to operate on point clouds one at a time and do not incorporate information from new scans in an incremental fashion. In particular, semantic segmentation is usually performed on individual scans or performed offline after complete scan data is collected. To overcome these shortcomings, this study proposes a multi-view incremental segmentation method that can perform online instance segmentation of 3D point clouds. The instance segmentation process assigns a class label and instance label to each point in the scanned point cloud. The class label is computed directly by a neural network whereas the instance label is determined by agglomerative clustering. The segmentation results are progressively updated in a dynamic fashion as the robot traverses its environment and acquires laser scan data, allowing real-time object recognition and improved scene understanding for the robot. The proposed work uses a novel multi-view context pooling (MCP) module to aggregate information from previous views before performing instance segmentation. The key idea of this work is to use contextual information from not all, but a select few of the previous scans, to improve segmentation of the current scan. This online incremental segmentation framework allows new scans to be processed within tenths of a second compared to half a minute if segmentation of the entire point cloud were to be recomputed with an offline method.

In summary, the main contributions of this work are as follows: (i) we develop a multi-view context pooling (MCP) module to combine information from multiple viewpoints (ii) we propose a joint network for online point cloud clustering and classification under a single framework, and (iii) we introduce a simulated ray-tracing dataset for labeled dynamic laser scan data \cite{github18}.

\section{RELATED WORK}

Semantic processing of point cloud data acquired from a mobile robot can be carried out in several forms such as clustering and classification. The clustering process aims to subdivide a large point cloud into smaller chunks that form semantically cohesive units. Clustering methods usually rely on generating seed points, then performing a region growing procedure to create a point cloud cluster around the seed point \cite{dube18}. Cluster assignment is propagated to neighboring points based on criteria such as distance, similarity of normal vectors, and similarity of color \cite{finman14}. 
More sophisticated methods for clustering use deep learning frameworks to infer a point feature embedding that can be used to predict point grouping proposals, for example, Similarity Group Proposal Networks (SGPN) \cite{wangweiyue18}. In this case, the distance between points in the learned embedding space is used as the criteria for grouping points together.

On the other hand, classification of point cloud data can be carried out at the scale of the point level, such that each point has an individual label. Some methods project the 3D point cloud into a 2D form and perform segmentation on the resulting image \cite{wu18}\cite{wang18}. Other methods use 3D convolutions to operate on a voxel grid and compute features in a layered fashion \cite{maturana15}\cite{riegler17}.
Due to the poor computational scalability of 3D convolutions, an alternative is to compute both point features and scene features derived from pooling operations, which are then concatenated to predict class probabilities for each point \cite{qi2017}\cite{qi17}.
Further advancements to this line of work use superpoint graphs \cite{landrieu18}, recurrent networks  \cite{engelmann17}, or coarse-to-fine hierarchies \cite{dai18} to incorporate neighborhood information and local dependencies to the prediction stage. 
However, these methods are usually applied in the offline setting, i.e. as a post-processing step after complete point cloud data is obtained, and do not take into account occlusion effects that comes into play with robotic scanning.

In contrast, for robotics applications, point cloud data of a scene is usually incrementally obtained in separate scans as the robot moves to different points around the site of interest. To make use of multi-view information, features from multiple viewpoints can be combined using operations such as global view pooling \cite{su15}, grouped pooling \cite{feng18}, or joint viewpoint prediction and categorization \cite{kanezaki18}. However, these methods perform classification at the object level in the offline setting, where the views are combined with complete point cloud information. 
For classification at the point level, several works \cite{antonello18}\cite{tateno16} use a method where point features are computed repeatedly for each observation and merged to determine the final classification. However, the final view merging process is still performed offline using computationally-heavy methods such as conditional random fields. 

In summary, conventional methods for semantic processing of point cloud data are either performed on individual scans or performed offline. The proposed work addresses these limitations by using an incremental method to incorporate new scan information in the instance segmentation task as opposed to considering scans on an individual basis.

\section{METHODOLOGY}

\subsection{3D scan data staging}

The Stanford 3D Indoor Spaces (S3DIS) dataset \cite{armeni16} is used to generate training and test data in the form of labeled 3D point clouds for this study. The dataset contains 6 building areas with point clouds segments that are individually annotated. It also contains 13 classes of objects which includes both building components such as walls and doors as well as furniture such as tables and chairs.
A virtual robot with laser scanners is placed in the environment and acquires laser scans by ray-tracing. The virtual robot is manually navigated in the building to ensure that all objects in the environment are adequately scanned. The robot is also assumed to have cameras so that color information can be mapped onto the point cloud.
The building environment is represented as a voxel grid and points are scanned whenever a laser ray intersects an occupied voxel. Note that the voxel grid is only used as a look-up table and the original point cloud coordinates (before rounding to the nearest voxel) are used for segmentation.
The robot traverses a pre-specified trajectory across the environment and dynamically acquires a scan every 0.2m. At each scan point, ray-tracing is carried out $360^{\circ}$ horizontally and $180^{\circ}$ vertically around the robot with a horizontal and vertical resolution of $1^{\circ}$. This can be emulated in real robots by physically rotating a planar laser scanner using a motor and ensures that the ceiling and floor area around the robot can be scanned.

\begin{figure}[thpb]
  \centering
	\subfloat[Original real point cloud of an indoor environment]{ \includegraphics[clip,width=0.9\linewidth]{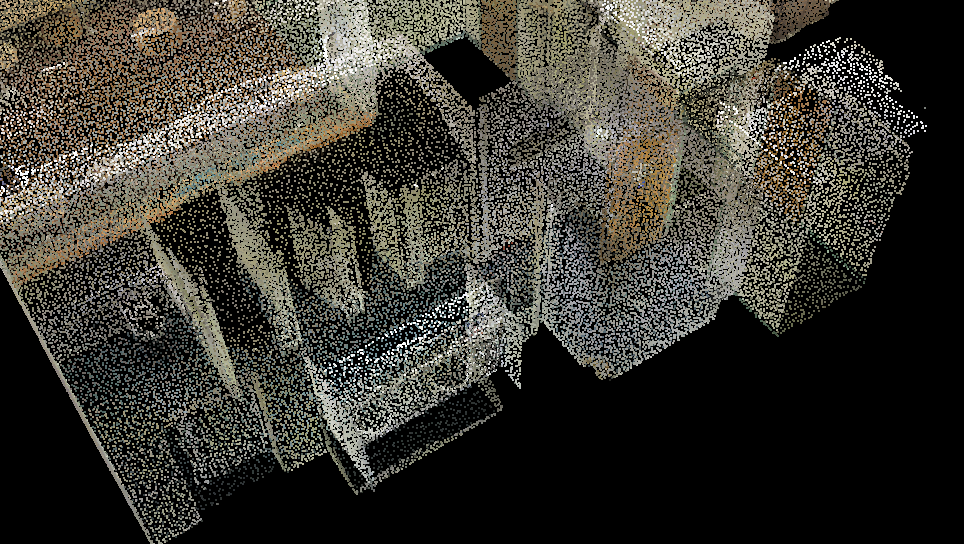} }
	\newline
	\subfloat[Synthetic point cloud from ray-tracing results along a trajectory]{ \includegraphics[clip,width=0.9\linewidth]{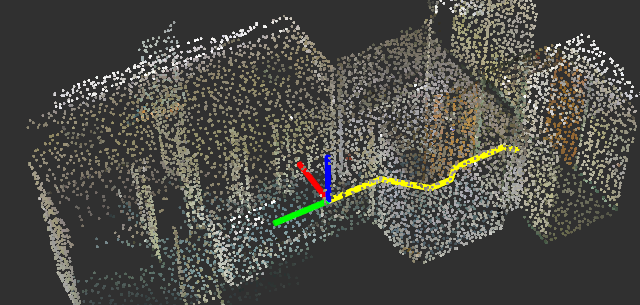} }
  \caption{Visualization of dynamic scanning by ray-tracing for a virtual mobile robot}
  \label{fig:raytrace}
\end{figure}

The ray-tracing process allows the acquired data to fully-capture the effects of occlusion and clutter that are common in real-world environments. 
Figure \ref{fig:raytrace} shows the original point cloud of a room as well as the resulting point cloud obtained by ray-tracing along a trajectory. This process assumes that the robot is capable of accurate localization such that scan data can be registered automatically. The ray tracing code is made publicly available \cite{github18}.

\subsection{Incremental point classification}

\begin{figure*}[t]
  \centering
  \includegraphics[width=0.9\linewidth,height=9cm]{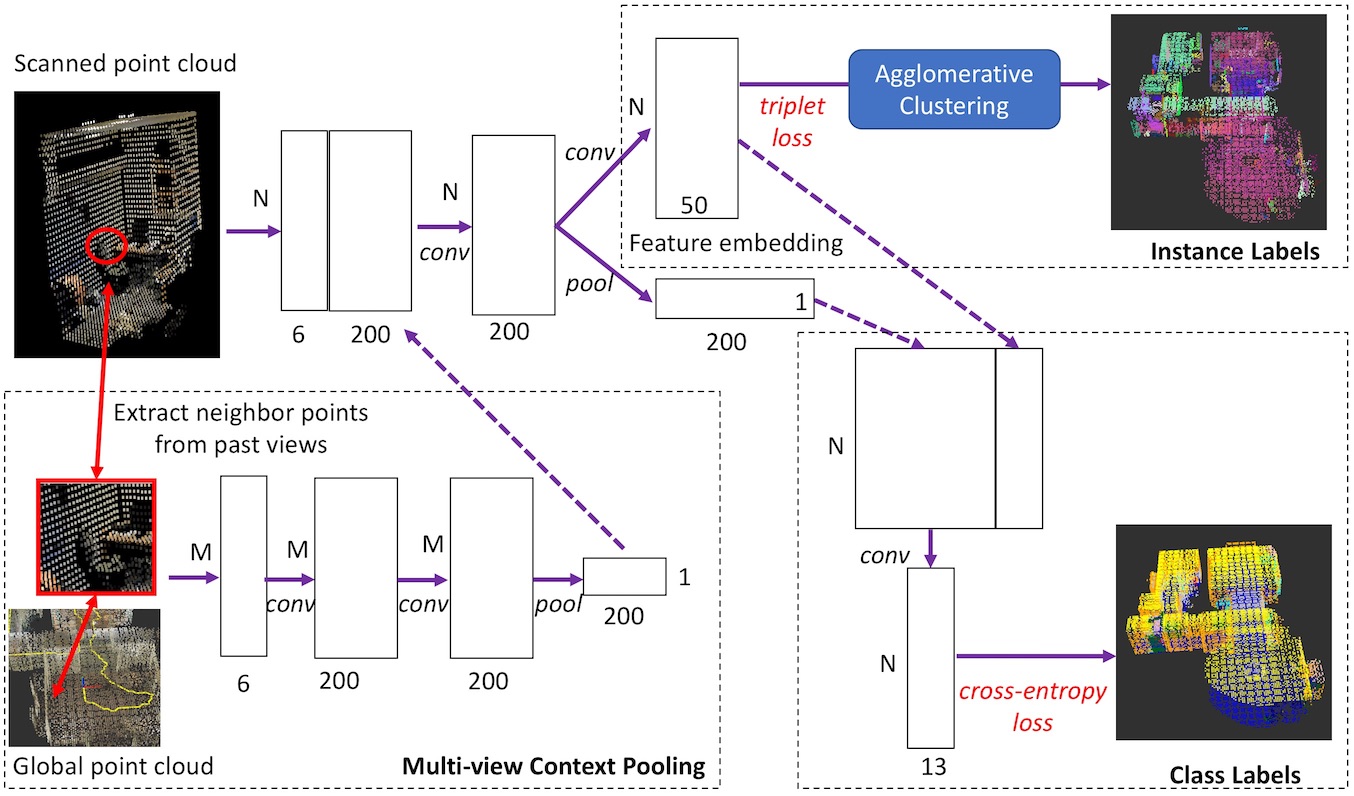}
  \caption{MCPNet architecture for incremental instance segmentation of 3D point clouds. The network consist of two branches: class labelling, described in Section IIIB, and instance labelling, described in Section IIIC}
  \label{fig:architecture}
\end{figure*}

The proposed incremental segmentation method involves a neural network architecture, named Multi-view Context Pooling Network (MCPNet), that jointly predicts an instance label and a class label for each scanned point. The class label describes the semantic class of a point (i.e. \textit{wall, door, table, etc.}) whereas the instance label is a unique ID such that only points from the same object instance have the same ID.

In order to perform instance segmentation on an online setting, it is important to keep track of newly scanned points and their relationship to previously scanned points in a global coordinate system. Thus, a voxel grid with grid size of 0.1m is used as a lookup-table to store the point coordinates and their corresponding segmentation results. The voxel grid ensures that points are only added to the lookup-table when new regions are scanned and prevents the data from growing too quickly, since the point cloud is limited to one point per voxel. The voxel grid is also key to efficiently retrieving neighbor points and context points for segmentation.

Each input scan acquired by the robot is first pre-processed to retain points only in an area of radius 2m around the robot. This is due to the fact that scan regions that are too far away from the robot are too sparse to make accurate predictions about the object class. Thus, faraway points are discarded for the current scan but can still be included in a future scan as the robot moves to a closer locations. Next, the point coordinates are normalized so that the x-y coordinates are centered around the robot and the z-coordinate is zero at the floor level. The points are then divided into batches of $N$x6 matrices, where $N$ is the batch size and the columns are X-Y-Z-R-G-B values. If there are more than $N$ points in the current scan, data will be passed to the network in multiple batches, where the last batch is sampled with replacement.

Figure \ref{fig:architecture} shows an outline of the proposed MCPNet architecture used to process the resulting input matrices. The input matrix is first passed through a 1D convolution layer to form an intermediate $N$x200 feature matrix.
The network then splits into two branches: the lower branch for classification and the upper branch for clustering, which will be described in detail in the next subsection. The lower branch uses a max pooling function to compute a global feature vector representing the input points, similar to PointNet\cite{qi17}. This is then concatenated with the feature embedding from the upper branch and passed through another 1D convolution layer to compute class probabilities for each point, represented as a $N$x13 matrix since there are 13 classes.

The proposed architecture also incorporates a Multi-view Context Pooling (MCP) module which incorporates contextual information from previous views to points in the current scan to improve the classification accuracy. The input to the MCP module is an $N$x$M$x6 tensor that stores context points for each of the $N$ points in the current input. A context point for point $p_i$ is defined as any point from previous views that is at most three cells away from $p_i$ in the global voxel grid. The three-cell distance threshold is set so that only points from previous views that are in close proximity to the current view are used as context. $M$ points are thus randomly sampled from among all the context points accumulated from past scans. The input tensor is passed through two layers of 1D convolutions as well as a max pooling layer. A $N$x200 matrix forms the output of the MCP module and is concatenated to the $N$x6 input matrix of the main network.

\begin{figure*}[t]
  \centering
  \includegraphics[width=0.9\linewidth]{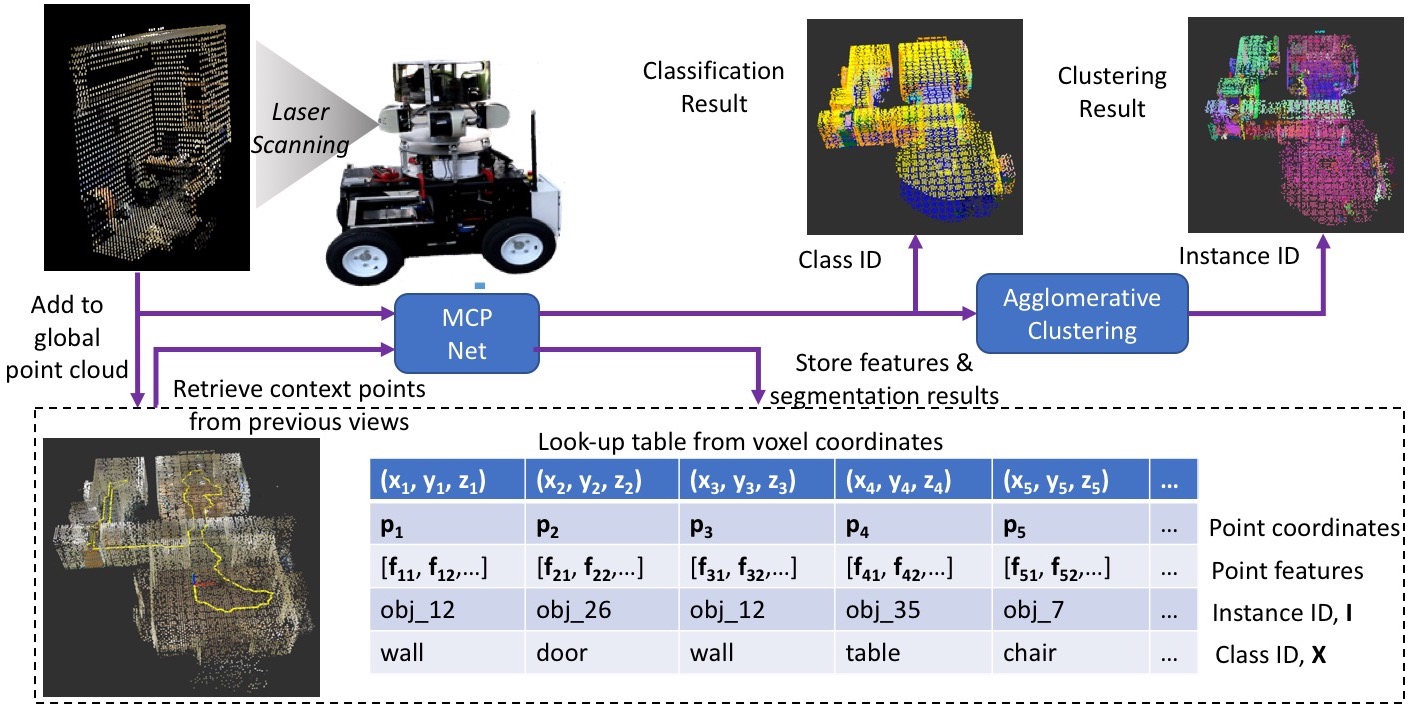}
  \caption{Overall framework for multi-view incremental segmentation}
  \label{fig:pipeline}
\end{figure*}

\subsection{Incremental clustering}

The upper branch of the proposed network (Figure \ref{fig:architecture}) aims to project the point features to an embedding space that is suitable for clustering, i.e. points from the same object should be close together whereas points from different objects should be further apart in the embedding space. 
This is enforced by computing a triplet loss \cite{schroff15} between sets of three points, $p_1,p_2$ which originate from the same object and $p_3$ which originates from a different object, and minimizing the resulting loss function.
Let the projection to the embedding space be represented as a function, $f:\mathbb{R}^6 \mapsto \mathbb{R}^{50}$. At training time, minimizing the triplet loss encourages the distance between points that should belong to the same object to be greater than that of points that should belong to different objects in the embedding space, i.e. $||f(p_1)-f(p_2)||^2 + \alpha < ||f(p_1)-f(p_3)||^2$, where $\alpha$ is the margin parameter.

At inference time, the instance labels for a new point cloud scan are determined by an agglomerative clustering scheme. First, a set of neighbor points are retrieved for each new point $p_i$. A neighbor point to $p_i$ is defined as a point that is at most one cell away from $p_i$ in the global voxel grid. Then, each point $p_i$ is only connected to a neighboring point $p_j$ if the cosine similarity, $\frac{f(p_i) \cdot f(p_j)}{||f(p_i)||\:||f(p_j)||}$ is greater than  $\beta$, where $\beta$ is a preset threshold. The following rules are then applied:
\begin{itemize}
\item If no connections exist, the point is initialized as the seed for a new cluster with a new instance ID.
\item If connections exist to a single existing cluster, the point is added to that cluster and takes the corresponding instance ID.
\item If connections exist to multiple existing clusters, all connected clusters are merged and their instance IDs are updated accordingly.
\end{itemize}

\subsection{Network training}

Training data is first collected by carrying out virtual scanning of the S3DIS dataset as described in Section III A. The proposed network is then trained from scratch in the offline setting. The two-branches are jointly optimized during training: the point-wise classification branch is trained with the softmax cross entropy loss, whereas the point feature embedding branch is trained with the triplet semihard loss, as described in \cite{schroff15}. The network is implemented in Tensorflow and trained with the ADAM optimizer. Training is carried out for 100 epochs with a learning rate of 0.001 and batch size of $N=256$ points.

\subsection{Overall segmentation framework}

Figure \ref{fig:pipeline} summarizes the overall segmentation framework. Point cloud data is stored in a global look-up table that efficiently maps voxel coordinates to point coordinates and segmentation results. Each time the robot acquires a new laser scan, points in previously-unseen voxels are added as new entries to the look-up table. The current scan is first normalized and passed to the network in batches of $N$ points. Context points from both current and previous views are also retrieved and passed to the MCP module. The network then computes a feature embedding as well as classification scores for each input point. The network output is recombined across all batches of the current scan and the global look-up table is updated on a point-by-point basis. The computed feature embedding is used to cluster the new points together with previous object instances. Finally, the class label and instance label of points in the current scan are updated accordingly.

\section{RESULTS}

\begin{figure*}[t]
\minipage{0.25\textwidth} \includegraphics[width=0.99\linewidth]{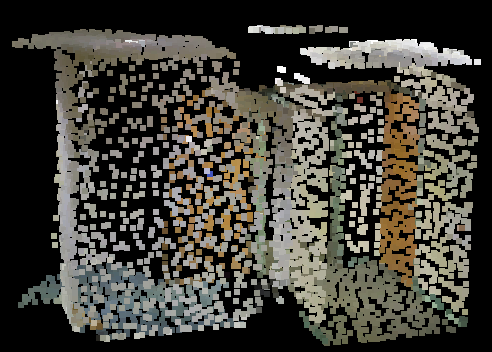} \endminipage\hfill
\minipage{0.25\textwidth} \includegraphics[width=0.99\linewidth]{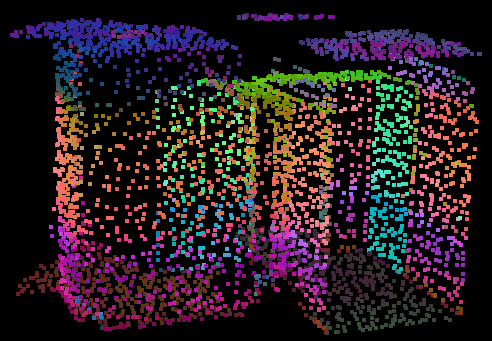} \endminipage\hfill
\minipage{0.25\textwidth} \includegraphics[width=0.99\linewidth]{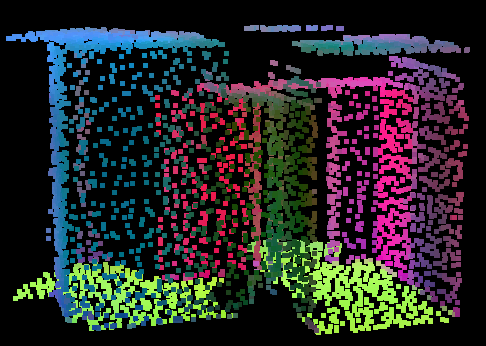} \endminipage\hfill
\minipage{0.25\textwidth} \includegraphics[width=0.99\linewidth]{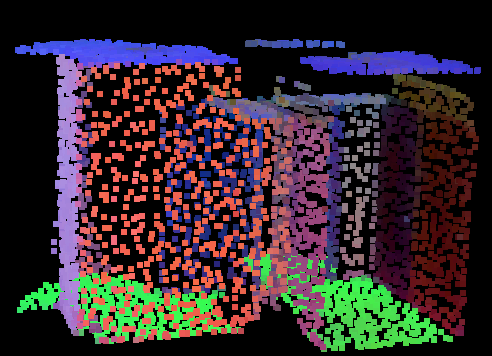} \endminipage
\caption{Visual comparison of feature embedding results: (i) original point cloud (ii) SGPN embedding (iii) proposed embedding (without MCP) (iv) proposed embedding (with MCP)} \label{fig:embedding}
\end{figure*}

\subsection{Evaluation of classification accuracy}
The instance segmentation results were first evaluated in terms of the classification accuracy on the S3DIS dataset \cite{armeni16}. A cross-validation scheme is used where one area of the dataset is held out as test data whereas the remaining areas are used as training data, repeated for 6 building areas. As shown in Table \ref{table:acc1}, the performance of the proposed method, with and without the MCP module, is compared against four other conventional point cloud classification methods, which were applied online on individual scans. PointNet, PointNet++, and VoxNet were trained on only the classification task whereas the remaining methods were trained on both the classification and clustering tasks. The result of applying PointNet and SGPN on an offline basis, that is on the complete point cloud after scanning, is also shown to obtain an upper bound for classification accuracy. Note that this baseline is obtained from the original post-scan dataset \cite{armeni16} which has significantly less occlusion and is additionally cleanly divided into rooms.

The evaluation metrics are: (i) Intersection-Over-Union (IOU), defined as the number of true positive points divided by the total true positive, false positive, and false negative points (according to the convention of \cite{landrieu18}\cite{engelmann17})
 (ii) point-wise accuracy, defined as the number of true positive points divided by total number of points (iii) object-wise accuracy, defined as the number of true positive object instances divided by total number of object instances. The IOU metric is additionally averaged over 13 classes. 
 The object-wise accuracy has lower values compared to point-wise accuracy since it requires an object to be both correctly segmented and have correctly classified points. 
 Experimental results show that the proposed method with MCP outperforms the other methods in terms of both IOU and accuracy.
\begin{table}[h]
  \caption{Classification accuracy on S3DIS dataset}
  \label{table:acc1}
  \centering
  {\renewcommand{\arraystretch}{1.0}
  \begin{tabular}{ccccc}
    \toprule
    Method & Avg IOU & Point Acc. & Obj. Acc.\\
    \midrule
    \shortstack{PointNet (offline)} & 49.8 & 79.9 & -\\
    \shortstack{SGPN (offline)} & 50.4 & 80.8 & -\\
    \midrule
PointNet \cite{qi2017} & 25.0 & 59.1 & 16.7 \\
PointNet++ \cite{qi17} & 23.8 & 60.2 & 14.1 \\
SGPN \cite{wangweiyue18} & 24.9 & 60.3 & 14.1 \\
VoxNet \cite{maturana15} & 18.2 & 45.5 & 6.5\\
    \midrule
Proposed & 25.4 & 57.7 & 15.6\\
Proposed + MCP & 39.2 & 74.9 & 33.5\\
    \bottomrule
  \end{tabular}}
\end{table} 

\begin{figure}[h]
  \centering
  \includegraphics[width=0.9\linewidth]{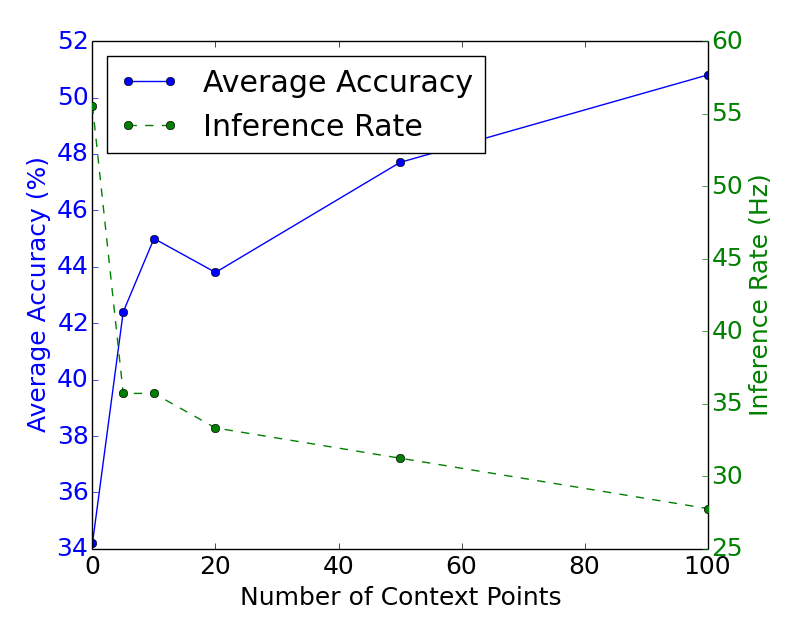}
  \caption{Classification accuracy and inference rate as a function of the number of context points}
  \label{fig:graph}
\end{figure}

Figure \ref{fig:graph} shows an analysis on the effect of the number of context points used in the proposed MCP module (Figure \ref{fig:architecture}) on the classification accuracy (trained on a smaller dataset). The graph shows that the average accuracy increases with number of context points but also incurs a higher computational cost. In the implementation used in this study, a total of 50 context points are used.

\subsection{Evaluation of clustering performance}

On the other hand, the clustering performance was measured using three different metrics: normalized mutual information (NMI), adjusted mutual information (AMI), and adjusted rand index (ARI), as defined in \cite{vinh10}. Table \ref{table:acc2} shows a comparison of the clustering metrics using different methods, with the region-growing scheme\cite{dube18}\cite{finman14} based on differences in normal vectors and color shown as a baseline.
In this section, PointNet, PointNet++, and VoxNet used the clustering technique proposed in \cite{qi2017}, where connected components are formed between points with the same class label. On the other hand, the remaining methods used the agglomerative clustering technique described in Section III.C, with $\beta=0.98$ for SGPN and $\beta=0.9$ for MCPNet (tuned by cross-validation). Results show that even the simple normal + color-based region-growing scheme can achieve a good clustering, but the proposed method with MCP achieved the best clustering result overall.

To validate the ability of the proposed network to project point features to an embedding space that is effective for clustering, the feature embedding is visualized as follows: (i) the point cloud is passed through MCPNet to compute a high-dimensional feature vector for each point (ii) Principal Component Analysis is used to reduce the feature vector to 3 dimensions (iii) the RGB color channels of each point is set to their corresponding 3-dimensional feature vector. Visual results in Figure \ref{fig:embedding} shows that the proposed method was able to produce distinct separation between points from different objects.

\begin{figure*}[t]
  \centering
  \includegraphics[width=0.99\linewidth]{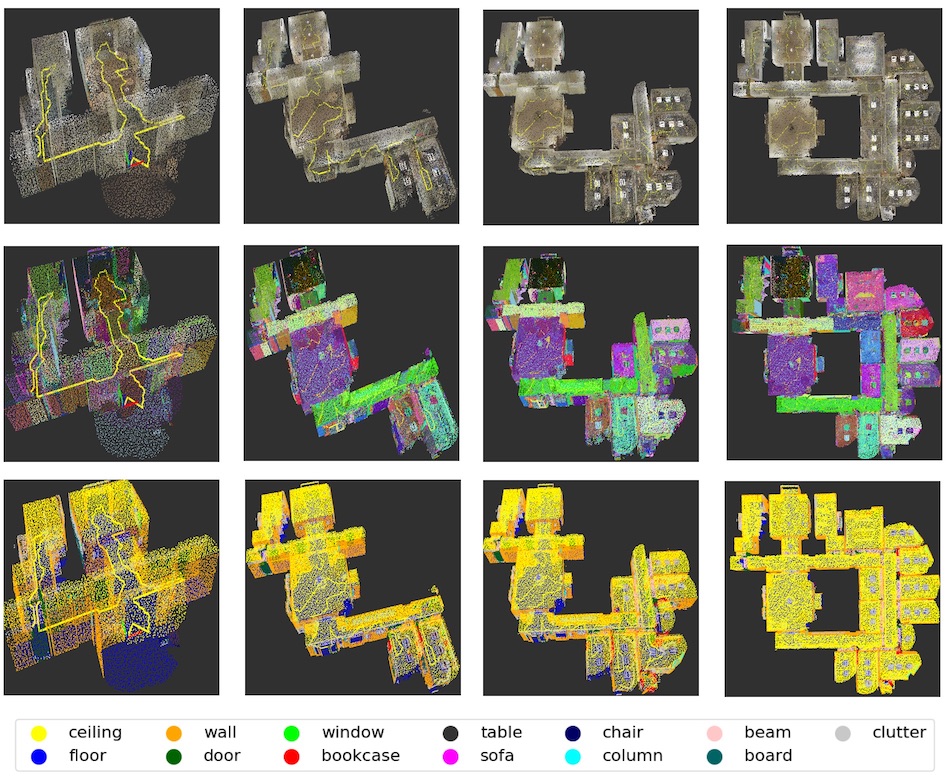}
  \caption{Incremental segmentation results on S3DIS test dataset: input point cloud with virtual robot trajectory (top row), clustering results (middle row) and classification results (bottom row). Color legend is shown for classification results.}
  \label{fig:ros}
\end{figure*}

\begin{figure*}[t]
  \centering
  \includegraphics[width=0.99\linewidth]{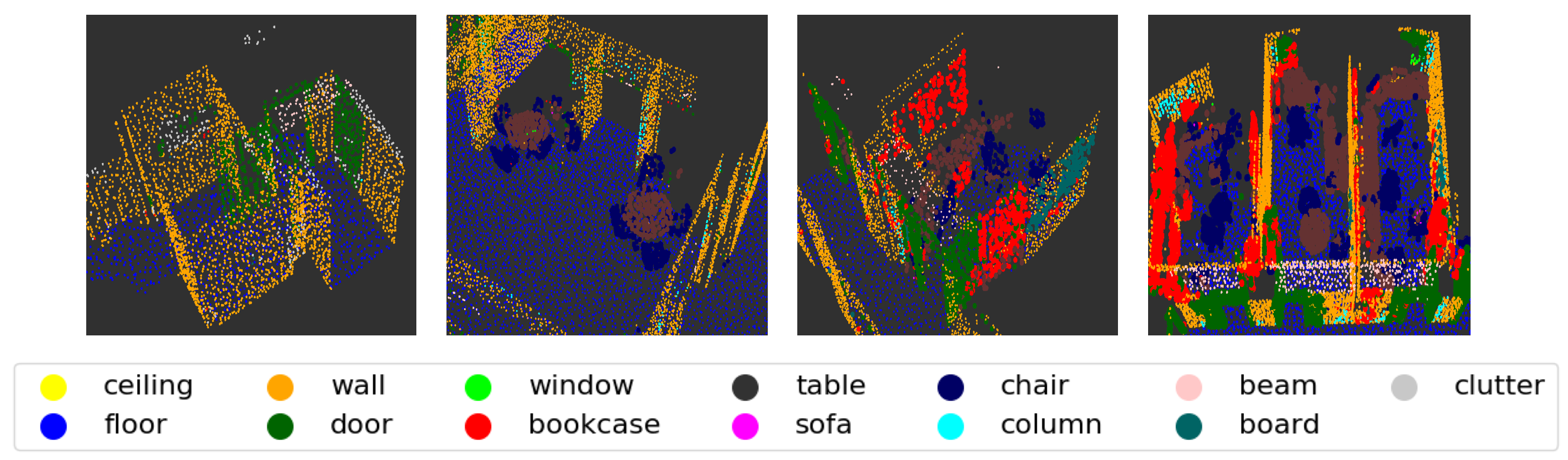}
  \caption{Close-up visualization of classification results with ceiling removed: (from left to right) (i) hallway with doors (ii) hallway with tables and chairs (iii) office room with board, bookcases, tables and chairs (iv) three adjacent offices}
  \label{fig:closeup}
\end{figure*}

\begin{table}[h]
  \caption{Clustering performance on S3DIS dataset}
  \label{table:acc2}
  \centering
  \begin{tabular}{cccc}
    \toprule
    Method & NMI & AMI & ARI\\
    \midrule
Normal + color & 78.5 & 63.4 & 25.0 \\
\midrule
PointNet \cite{qi2017} & 72.1 & 59.2 & 12.1 \\
PointNet++ \cite{qi17} & 76.1 & 66.4 & 17.1 \\
SGPN \cite{wangweiyue18} & 78.5 & 60.5 & 26.1 \\
VoxNet \cite{maturana15} & 70.5 & 58.0 & 11.3 \\
    \midrule
Proposed & 75.9 & 64.8 & 18.3 \\
Proposed + MCP & 85.6 & 74.2 & 39.7 \\
    \bottomrule
  \end{tabular}
\end{table} 

Finally, Figure \ref{fig:ros} shows the global instance segmentation results at intermediate points along the incremental scanning process whereas Figure \ref{fig:closeup} shows close-up views of specific regions in the scanned environment.

\subsection{Efficiency evaluation}

Table \ref{table:perf1} shows the computation time and memory usage of each segmentation method, measured on an Intel Xeon E3-1200 CPU with a NVIDIA GTX1080 GPU. Results show that the proposed method with MCP led to an increase in processing time per scan, but at a reasonable amount such that it can still be run in real time. Whereas, the memory usage is similar in magnitude to competing methods. 

\begin{table}[h]
  \caption{Efficiency comparison for instance segmentation}
  \label{table:perf1}
  \centering
  {\renewcommand{\arraystretch}{1.0}
  \begin{tabular}{cccc}
    \toprule
    Method & \shortstack{Processing Time\\per Scan (s)} & \shortstack{Network\\Size (MB)} & \shortstack{CPU Memory\\Usage (GB)}\\
    \midrule
PointNet \cite{qi2017} & 0.017 & 14.0 & 0.94 \\
PointNet++ \cite{qi17} & 0.027 & 11.5 & 0.97 \\
SGPN \cite{wangweiyue18} & 0.019 & 14.9 & 2.49 \\
VoxNet \cite{maturana15} & 0.017 & 0.5 & 1.22 \\
\midrule
Proposed & 0.019 & 0.9 & 1.26 \\
Proposed + MCP & 0.034 & 1.8 & 1.26 \\
    \bottomrule
  \end{tabular}}
\end{table}

\section{CONCLUSIONS}

This study demonstrated a framework for performing incremental instance segmentation of laser-scanned point cloud data on a mobile robot using multi-view contextual information. A method for simulating dynamic scan data by ray-tracing building models was introduced. This ray-traced dataset, derived from the Stanford 3D Indoor Spaces (S3DIS) dataset, was used to train and test a deep neural network for instance segmentation and we release the code to allow reproduction of the dataset. Experimental results showed that the proposed approach led to 15\% improvement in accuracy and 7\% improvement in NMI compared to the next best online method.


\section*{ACKNOWLEDGMENT}

The work reported herein was supported by the United States Air Force Office of Scientific Research (Award No. FA2386-17-1-4655) and by a grant (18CTAP-C144787-01) funded by the Ministry of Land, Infrastructure, and Transport (MOLIT) of Korea Agency for Infrastructure Technology Advancement (KAIA). Any opinions, findings, and conclusions or recommendations expressed in this material are those of the authors and do not necessarily reflect the views of the United States Air Force and MOLIT. Zsolt Kira was partially supported by the National Science Foundation and National Robotics Initiative (grant \# IIS-1426998).


{\small
\bibliographystyle{IEEEtran}
\bibliography{main}
}

\end{document}